\documentclass[10pt,twocolumn,letterpaper]{article}
\usepackage[accsupp]{axessibility}
\usepackage{cvpr}              
\definecolor{cvprblue}{rgb}{0.21,0.49,0.74}
\usepackage[pagebackref,breaklinks,colorlinks,allcolors=cvprblue]{hyperref}

\usepackage[most]{tcolorbox}

\tcbset{
  promptstyle/.style={
    enhanced,                 
    breakable,                
    colback=gray!5!white,     
    colframe=gray!40!white,   
    boxrule=0.3pt,            
    arc=2pt,                  
    left=6pt,right=6pt,top=4pt,bottom=4pt, 
    before skip=10pt,after skip=10pt,      
    fonttitle=\bfseries,      
    title={Prompt},           
    coltitle=black,           
  },
}

\def\paperID{XXXXX} 
\def\confName{CVPR}
\def\confYear{2026}

\title{Tokenization Allows Multimodal Large Language Models to Understand, Generate and Edit Architectural Floor Plans}

\author{
    Sizhong Qin$^{1,2}$ \quad 
    Ramon Elias Weber$^{\dag 2}$ \quad 
    Xinzheng Lu$^{\dag 1}$ \\
    $^1$Tsinghua University \quad $^2$UC Berkeley \\
    {\tt\small \url{https://housemind.github.io/}} \quad {\small $^{\dag}$equal contribution} 
}

\begin{document}
\twocolumn[{%
\renewcommand\twocolumn[1][]{#1}%
\maketitle
\begin{center}
    \vspace{-2.5em}
    \captionsetup{type=figure}
        \includegraphics[width=0.9\textwidth]{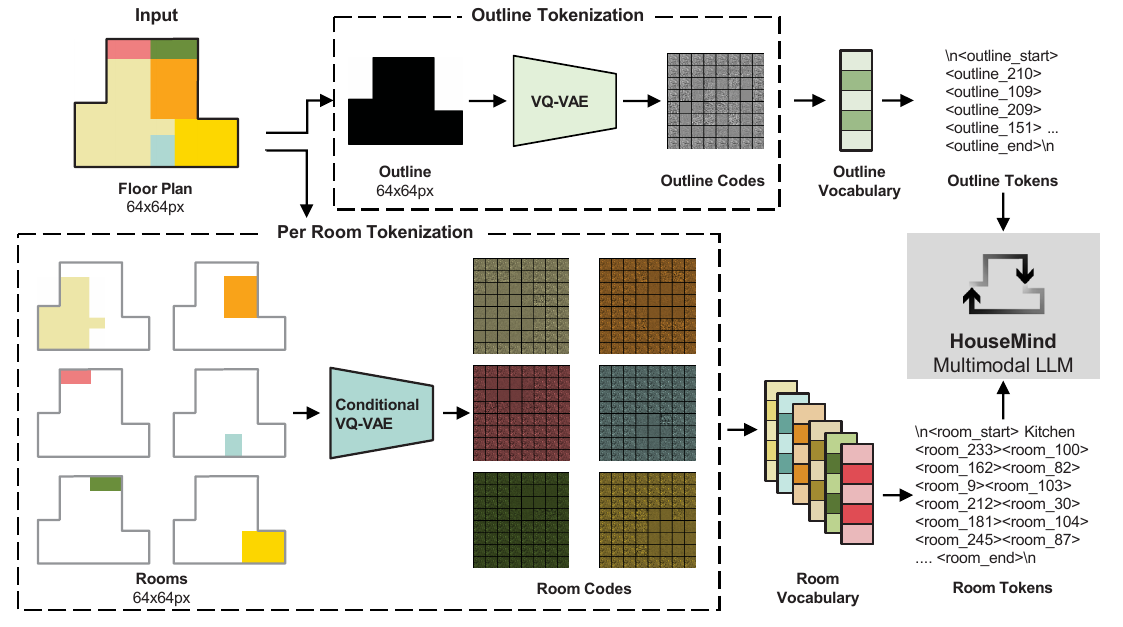}
    \\
    \vspace{-0.5em}
    \captionof{figure}{\textbf{HouseMind} learns the language of space by modeling outlines and rooms as spatial tokens. 
Through hierarchical tokenization and multimodal reasoning, it can understand, generate, and edit architectural floor plans from natural language prompts.
}
    \label{fig:teaser}
\end{center}
}]
\begin{abstract}
Architectural floor plan design demands joint reasoning over geometry, semantics, and spatial hierarchy, which remains a major challenge for current AI systems. Although recent diffusion and language models improve visual fidelity, they still struggle with coherent spatial reasoning and controllable generation. We present HouseMind, a multimodal large language model that unifies floor plan understanding, generation, and editing in one framework. We introduce discrete room-instance tokens to construct a unified vocabulary that bridges layouts and symbolic reasoning. With multimodal alignment and instruction tuning, the model synthesizes coherent, controllable layouts from text instructions. Experiments show how the framework achieves superior geometric validity and controllability while remaining efficient and locally deployable.
\end{abstract}    
\section{Introduction}
\label{sec:intro}
Generative models and artificial intelligence (AI) driven workflows are reshaping how buildings are designed~\cite{jangGenerativeAIArchitectural2025a, khanGenerativeAIApproaches2025, liaoGenerativeAIDesign2024a, zhuangMachineLearningGenerative2025}. They have been applied at different stages of the design process, including schematic generation, interactive co-design, and performance optimization~\cite{savovGeneralistGenerativeAgent2025, qinIntelligentDesignOptimization2024}. Although large language models (LLM) have been proven to be exceptional in generalizing, understanding, and emulating language, code, and pixels, they currently lack the ability for semantic reasoning that is required for spatial design. For the design of buildings, specifically floor plans, patterns are not sequential, but embedded in complex relationships. Due to this, the generation of an architectural layout remains one of the most cognitively demanding tasks~\cite{weberAutomatedFloorplanGeneration2022, yanGenerativeDesignArchitectural}. It requires models to capture hierarchical and relational dependencies between functional spaces while maintaining geometric feasibility and semantic coherence. In this research, we propose a new model that decodes spatial relationships and enables spatial coherence from prompt to output.

Recent advances in diffusion-based and autoregressive models have greatly improved the fidelity and diversity of layout generation.  However, these approaches still face four major limitations. First, they often regard layout synthesis as a purely visual process, without explicit reasoning at the room-instance level, which can lead to plans that appear locally plausible but lack global spatial coherence, such as consistent adjacency or circulation relationships among rooms. Moreover, preference-aligned or large-scale vision–language models frequently behave as black-box generators, offering limited interpretability and spatial controllability. At the same time, existing frameworks struggle to unify understanding, generation, and editing within a single architecture, particularly under the geometric and semantic complexity of building layouts. In addition, most AI systems remain computationally demanding and difficult to deploy locally, restricting their integration into practical design workflows.

In this work, we introduce \textbf{HouseMind}, an efficient and locally deployable multimodal model that unifies floor plan understanding, generation, and editing within one coherent framework. 
\textbf{HouseMind} discretizes layouts into room-instance tokens using a Vector-Quantized Variational Autoencoder (VQ-VAE)~\cite{vqvae} and leverages an LLM for multimodal reasoning. 
By representing both geometry and semantics as discrete token sequences, \textbf{HouseMind} bridges the gap between symbolic reasoning and continuous layout geometry, enabling controllable and interpretable operations directly in the latent token space.

Our approach offers three key advantages: (1) Fine-grained reasoning at the room-instance level allows text-guided control over spatial structure and semantic composition. (2) Unified multitask formulation: The same model handles understanding, conditional generation, and localized editing within a single sequence modeling framework. (3) Practical efficiency: The compact architecture enables real-time inference and on-device deployment while maintaining global coherence. To evaluate this unified formulation, we construct a benchmark encompassing all three tasks (understanding,  generation, and editing) based on the RPLAN dataset~\cite{wuDatadrivenInteriorPlan2019}. 
Extensive experiments demonstrate that \textbf{HouseMind} achieves strong generalization across modalities and tasks, outperforming prior diffusion- and LLM-based baselines while maintaining geometric validity, semantic consistency, and practical deployability.

\section{Related Work}
\label{sec:related_work}

\noindent\textbf{Learning spatial and functional patterns.}
Early data-driven approaches learn geometric and functional regularities of architectural layouts directly from data. 
GAN-based frameworks~\cite{luoFloorplanGANVectorResidential2022,nauataHouseGANRelationalGenerative2020,nauataHouseGANGenerativeAdversarial2021,aalaeiArchitecturalLayoutGeneration2023a,yeGraphRWGANMethodGenerating2025,liuIntelligentFloorPlan2024,tangGraphTransformerGANs2023}
enhance realism through adversarial and graph-constrained objectives, but often overfit to local geometries and lack global spatial semantics. 
Graph- or GNN-based approaches~\cite{huGraph2PlanLearningFloorplan2020,luComplexLayoutGeneration2025,xiaInteractiveAIGenerative2024,xinPromptsLayoutsHybrid2025,duptyConstrainedLayoutGeneration2024a,weberHypergraphModelShows2024a}
model room connectivity and hierarchy to improve relational reasoning, though discrete graph representations limit geometric fidelity and scalability.
Diffusion-based methods~\cite{suFloorPlanGraph2024,zengResidentialFloorPlans2024,guezeFloorPlanReconstruction2023,huGSDiffSynthesizingVector2025,zhangGeneratingAccessibleMultioccupancy2025,wangEliminatingRasterizationDirect2025}
generate diverse and stable results via iterative denoising, yet remain computationally costly and typically confined to single-task synthesis without high-level control.

\noindent\textbf{Incorporating structural and semantic reasoning.}
To go beyond pure pattern learning, recent studies integrate explicit structural constraints and semantic control into generative frameworks.
Structure-aware representations such as wall graphs or hierarchical layouts~\cite{huGSDiffSynthesizingVector2025,sunWallPlanSynthesizingFloorplans2022a} enable relationally consistent generation, while reinforcement and multi-agent frameworks~\cite{suGenerativeDesignComplex2024,zhouAutomatedAggregationDwelling2025,luoControllableFlexibleResidential2025,kakooeeEnhancingArchitecturalSpace2025} optimize functional objectives and circulation logic.
Further bridging geometry and semantics, MaskPLAN~\cite{zhangMaskPLANMaskedGenerative2024} introduces a VQ-VAE–based attribute discrete latent model that encodes geometric attributes into visual tokens and reconstructs them via masked transformer autoencoding, offering an early attempt at controllable semantic generation.
More recently, hybrid transformer--diffusion paradigms couple geometric decoding with structural reasoning for improved controllability~\cite{guezeFloorPlanReconstruction2023,zengResidentialFloorPlans2024}.
Despite these advances, existing methods still rely on handcrafted priors or task-specific constraints, limiting their generalization across diverse design contexts.

\noindent\textbf{LLM-driven multimodal design.}
Recent progress in LLMs and multimodal large models (MLLMs) introduces a new paradigm that connects textual intent with spatial reasoning. 
Tell2Design~\cite{lengTell2DesignDatasetLanguageGuided2023} establishes a benchmark linking textual descriptions and floor plan layouts.
ChatDesign~\cite{liChatDesignBootstrappingGenerative2024a} and DStruct2Design~\cite{luo2024dstruct2design} leverage LLM priors for layout synthesis. More recent frameworks such as LLM-based FloorPlan Design~\cite{qiuLLMbasedFrameworkAutomated2025} 
leverage LLMs to translate natural language into vectorized floor plans via structured semantic parsing, 
while ChatHouseDiffusion~\cite{qin2024chathousediffusion}
integrates language understanding with diffusion-based generation for enhanced controllability. 
Cross-modal agents like CARD~\cite{zengCARDCrossmodalAgent2024a} and ~\citet{zengUnifiedResidentialFloor2025} unify generation and editing, 
and FloorPlan-LLaMa~\cite{yinFloorPlanLLaMaAligningArchitects2025} and FloorPlan-DeepSeek~\cite{yin2025floorplandeepseek}
align multimodal reasoning with expert feedback for semantic understanding and next-room prediction.
Overall, while these efforts greatly enhance interpretability and cross-task reasoning, most models remain modular. 
This motivates our unified multitask multimodal framework that jointly learns geometric, semantic, and topological representations for consistent reasoning across understanding, generation, and editing.

\section{Problem Formulation}
\label{sec:formulation}

An architectural floor plan with $N$ rooms can be decomposed into two components:
an \textbf{outline} $x_o$ that defines the global floor plan boundary,
and a set of \textbf{room instances} $\{x_{r_i}\}_{i=1}^{N}$ that describe
the geometry and semantics of individual rooms:
\begin{equation}
x \;=\; \{\,x_o,\; \{x_{r_i}\}_{i=1}^{N}\,\}.
\end{equation}

\vspace{3pt}
\noindent\textbf{Discrete representation.}
Both the outline and the rooms are quantized by two separate VQ-VAE encoders:
\begin{equation}
\boldsymbol{z}_o \;=\; E_o(x_o), 
\qquad
\boldsymbol{z}_{r_i} \;=\; E_r(x_{r_i},\, x_o),
\end{equation}
where 
$\boldsymbol{z}_o = (z^{(o)}_{1},\dots,z^{(o)}_{m_o})$ and 
$\boldsymbol{z}_{r_i} = (z^{(r)}_{i,1},\dots,z^{(r)}_{i,m_i})$ 
denote the sequences of discrete tokens obtained from the outline and the $i$-th room, respectively.  
Here $m_o$ and $m_i$ are the numbers of tokens for the outline and room $r_i$,  
and $z^{(o)}_j\!\in\!\mathcal{Z}_o$, $z^{(r)}_{i,j}\!\in\!\mathcal{Z}_r$ come from the learned codebooks 
$\mathcal{Z}_o$ (outline) and $\mathcal{Z}_r$ (room).

\begin{figure}[htbp]
    \centering
    \includegraphics[width=\linewidth]{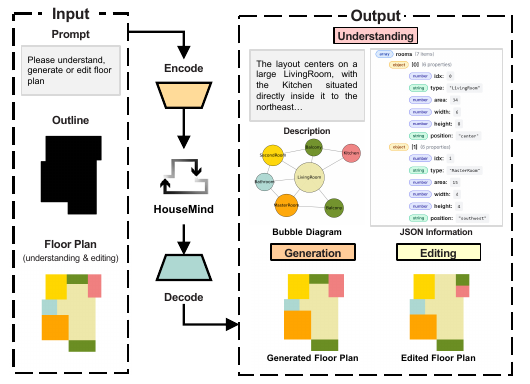}
    \caption{
    \textbf{Understanding}: given a prompt, an outline, and an existing floor plan, the model outputs a textual description, a bubble diagram, and structured JSON capturing spatial semantics.
    \textbf{Generation}: given a prompt and an outline, the model produces a complete, coherent floor plan.
    \textbf{Editing}: given a prompt, an outline, and a reference floor plan, the model outputs an updated plan aligned with the editing intent.
    }
    \label{fig:task}
\end{figure}

\vspace{3pt}
\noindent\textbf{Structured tokenization.}
The entire floor plan is represented as an interleaved token sequence combining geometric and semantic information:
\begin{equation}
Z \;=\; [\,\boldsymbol{z}_o,\; \ell_{r_1},\; \boldsymbol{z}_{r_1},\; \dots,\; \ell_{r_N},\; \boldsymbol{z}_{r_N}\,],
\end{equation}
where $\ell_{r_i}\!\in\!\mathcal{C}$ is the semantic label token of room $r_i$,  
and $\mathcal{C}$ denotes the set of all room categories.

\vspace{3pt}
\noindent\textbf{Unified task formulation.}
Given the structured sequence $Z$, the proposed \textbf{HouseMind} framework
jointly addresses three core tasks as shown in Fig.~\ref{fig:task}:

\begin{itemize}[leftmargin=1.2em]
    \item \textbf{Understanding:}
    inferring room functions, spatial relations, and topological constraints directly from $Z$.

    \item \textbf{Generation:}
    given a text specification $s$ (e.g., ``three bedrooms and one bathroom'')
    and outline tokens $\boldsymbol{z}_o$, the model autoregressively generates the layout:
    \begin{equation}
    p\!\left(Z \mid \boldsymbol{z}_o, s\right)
    \;=\;
    \prod_{t} p\!\big(Z_t \mid Z_{<t},\, \boldsymbol{z}_o,\, s\big),
    \end{equation}
    where $Z_t$ denotes the $t$-th token in the sequence.

    \item \textbf{Editing:}
    given an existing layout sequence $Z^{\mathrm{src}}$ and a text instruction $s$,
    the model produces an updated layout $Z^{\mathrm{tgt}}$ as
    \begin{equation}
    p\!\left(Z^{\mathrm{tgt}} \mid Z^{\mathrm{src}}, s\right)
    \;=\;
    \prod_{t} p\!\Big(Z^{\mathrm{tgt}}_t \,\Big|\, Z^{\mathrm{src}},\, Z^{\mathrm{tgt}}_{<t},\, s\Big),
    \end{equation}
    modifying only tokens relevant to the instruction while keeping others unchanged.
\end{itemize}

\vspace{3pt}
\noindent
In summary, \textbf{HouseMind} unifies floor plan understanding, generation, and editing as a single sequence-modeling problem over discretized outline and room tokens, enabling LLMs to jointly reason about spatial semantics and geometric structure.

\section{Method}
\label{sec:method}

Our framework, \textbf{HouseMind}, unifies geometric understanding, generation, and editing of architectural floor plans within a MLLM.
It consists of two core components: 
(1) \textbf{Room-Instance Tokenization}, which discretizes structural layouts into compact spatial tokens using hierarchical VQ-VAE modules; and 
(2) \textbf{Multimodal Alignment and Instruction Tuning}, which aligns spatial and linguistic representations for unified reasoning, generation, and editing.

\begin{figure*}[t]
    \centering
    \includegraphics[width=\linewidth]{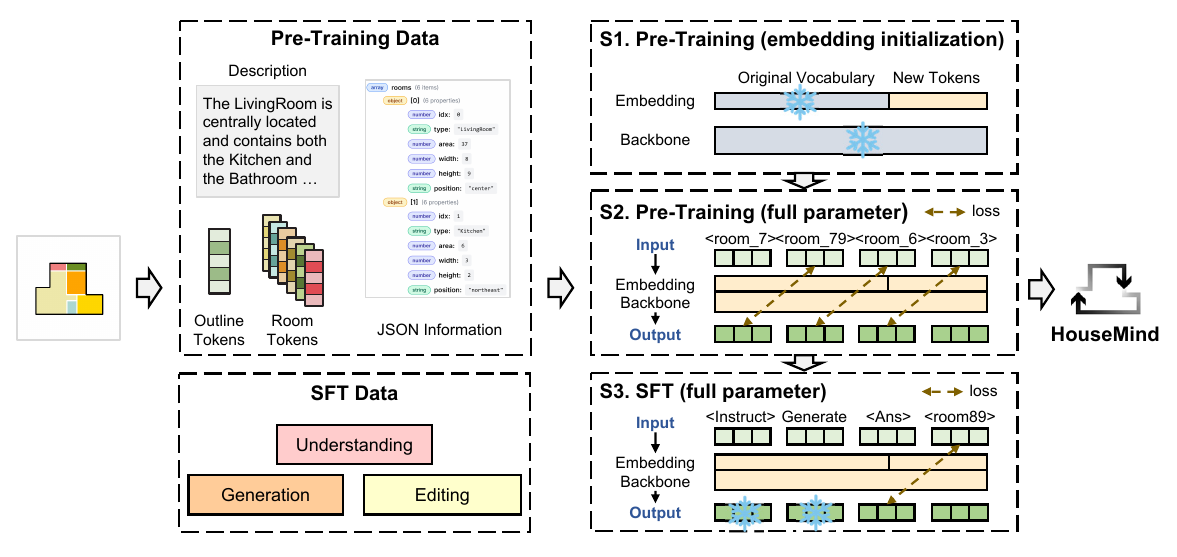}
    \caption{
    \textbf{Overall framework of HouseMind.}
    The model is trained through a three-stage multimodal alignment and instruction tuning pipeline:
    (S1) Embedding Initialization establishes cross-modal compatibility between geometric and linguistic tokens;
    (S2) Multimodal Pre-training aligns text and spatial representations;
    and (S3) Instruction Tuning (SFT) enables task-aware spatial reasoning.
    }
    \label{fig:housemind}
\end{figure*}

\subsection{Room-Instance Tokenization}
\label{subsec:tokenization}

To bridge continuous geometric layouts and discrete sequence modeling, we employ VQ-VAE modules to learn discrete representations of the floor plan outline and individual room instances. Each branch adopts an encoder, quantizer, and decoder structure with a CNN encoder, a transposed-CNN decoder, and its own learnable codebook.

\vspace{3pt}
\noindent\textbf{Outline discretization.}
A CNN encoder $E_o(\cdot)$ extracts latent features from the binary outline mask $x_o$. These features are vector-quantized using the outline codebook $\mathcal{Z}_o=\{e^{(o)}_k\}_{k=1}^{K_o}$:
\begin{equation}
z^{(o)}_j = e^{(o)}_{k_j^\star}, \quad
k_j^\star = \arg\min_{k}\!\big\lVert E_o(x_o)_j - e^{(o)}_{k}\big\rVert_2 .
\end{equation}
Here $j=1,\dots,m_o$ indexes the outline tokens.  
The decoder reconstructs the outline as $\hat{x}_o = D_o(\boldsymbol{z}_o)$, where $\boldsymbol{z}_o=(z^{(o)}_1,\dots,z^{(o)}_{m_o})$.  
This process converts geometric contours into a discrete vocabulary that captures the global building envelope.

\vspace{3pt}
\noindent\textbf{Conditional room discretization.}
Each room mask $x_{r_i}$ is encoded jointly with the corresponding outline context. The conditional encoder $E_r(\cdot)$ maps $(x_{r_i},x_o)$ to latent features, which are then quantized using the room codebook $\mathcal{Z}_r=\{e^{(r)}_k\}_{k=1}^{K_r}$:
\begin{equation}
z^{(r)}_{i,j} = e^{(r)}_{k_{i,j}^\star}, \quad
k_{i,j}^\star = \arg\min_{k}\!\big\lVert E_r(x_{r_i},x_o)_j - e^{(r)}_{k}\big\rVert_2 .
\end{equation}
Here $j=1,\dots,m_i$ enumerates tokens within room $r_i$.  
The decoder reconstructs each room as $\hat{x}_{r_i}=D_r(\boldsymbol{z}_{r_i},x_o)$, where $\boldsymbol{z}_{r_i}=(z^{(r)}_{i,1},\dots,z^{(r)}_{i,m_i})$.  
By conditioning room encoding on the outline, the model learns context-aware room representations that capture both geometry and spatial adjacency, forming a structured and interpretable spatial token sequence.

\subsection{Multimodal Alignment and Instruction Tuning}
\label{subsec:housemind}

Built upon discretized spatial tokens, HouseMind integrates language and geometry through a three-stage multimodal alignment and instruction tuning pipeline (Fig.~\ref{fig:housemind}) that progressively enhances spatial reasoning, cross-modal understanding, and controllable generation.

\vspace{3pt}
\noindent\textbf{Stage 1. Embedding initialization.}
Spatial codebooks obtained from the VQ-VAE modules, including the outline codes \(\mathcal{Z}_o\) and room codes \(\mathcal{Z}_r\), are incorporated into the language model’s vocabulary by assigning each code a unique trainable token embedding.  
This establishes a one-to-one correspondence between discrete spatial codes and the model’s textual tokens, forming a unified vocabulary that jointly represents geometry and language.  
Through this initialization, HouseMind ensures that spatial and linguistic symbols coexist within the same token space, allowing the MLLM to process geometric layouts and natural language seamlessly within a single autoregressive sequence.

\begin{table*}[t]
\centering
\caption{\textbf{Understanding results.}
Success: success rate; RMR: room match rate; LocAcc: room location accuracy; 
AreaDiff: room area difference (m$^2$); AdjAcc: room adjacency accuracy; RelAcc: spatial relation accuracy.}
\label{tab:understanding}
\setlength{\tabcolsep}{5pt}
\begin{tabular}{lccccccc}
\toprule
\textbf{Method} & \textbf{Success} & \textbf{RMR} & \textbf{LocAcc} & \textbf{AreaDiff$\downarrow$} & \textbf{AdjAcc} & \textbf{RelAcc} & \textbf{Time (s)} \\
\midrule
LLaVA-v1.6-Mistral-7B-HF & \textbf{1.000} & 0.616 & 0.225 & 3.649 & 0.134 & 0.056 & $\sim$6 \\
Qwen3-VL-8B-Instruct     & \textbf{1.000} & 0.698 & 0.347 & 5.837 & 0.382 & 0.128 & $\sim$8 \\
InternVL3.5-8B          & \textbf{1.000} & 0.847 & 0.546 & 12.234 & 0.469 & 0.157 & $\sim$13 \\
MiniCPM-V~4.5           & 0.996 & 0.904 & 0.492 & 13.765 & 0.597 & 0.208 & $\sim$14 \\
\midrule
HouseMind-U              & \textbf{1.000} & \textbf{0.998} & \textbf{0.969} & \textbf{0.549} & \textbf{0.990} & \textbf{0.808} & $\sim$3 \\
HouseMind-O           & \textbf{1.000} & \textbf{0.998} & 0.925 & 0.655 & 0.954 & 0.738 & $\sim$3 \\
\bottomrule
\end{tabular}
\end{table*}

\begin{table*}[t]
\centering
\caption{\textbf{Generation results.}
Micro/Macro IoU measure pixel-level overlap; 
SSIM~\cite{ssim} and PSNR quantify perceptual similarity;
FID~\cite{fid} and GED~\cite{ged} evaluate distributional realism;
Node~F1 and Edge~Overlap assess graph-level correctness.
* denotes methods without released code; results are reproduced.}
\label{tab:generation}
\setlength{\tabcolsep}{5pt}
\begin{tabular}{lccccccccc}
\toprule
\textbf{Method} &
\textbf{Micro IoU} & 
\textbf{Macro IoU} & 
\textbf{SSIM} & 
\textbf{PSNR} & 
\textbf{FID}$\downarrow$ & 
\textbf{GED}$\downarrow$ & 
\textbf{Node F1} & 
\textbf{Edge Ovl.} & 
\textbf{Time (s)} \\
\midrule
Qwen-Image-Edit-2509 & 0.161 & 0.0621 & 0.721 & 12.4 & 156 & -- & -- & -- & $\sim$240 \\
Tell2Design                  & 0.390 & 0.307  & 0.840 & 13.2 & 30.5 & 6.94 & 0.808 & 0.197 & $\sim$15 \\
ChatHouseDiffusion           & 0.589 & 0.521  & 0.866 & 14.9 & 11.3 & 2.36 & 0.985 & 0.710 & $\sim$30 \\
FloorPlanLLaMA*              & 0.607 & 0.511  & 0.874 & 15.5 & 49.3 & 2.68 & 0.922 & 0.574 & $\sim$1 \\
\midrule
HouseMind-G                & 0.709 & 0.653  & 0.886 & 16.0 & 1.91 & \textbf{1.01} & \textbf{0.994} & \textbf{0.880} & $\sim$2 \\
HouseMind-O             & \textbf{0.710} & \textbf{0.654} & \textbf{0.887} & \textbf{16.1} & \textbf{1.89} & 1.03 & \textbf{0.994} & \textbf{0.880} & $\sim$2 \\
\bottomrule
\end{tabular}
\end{table*}

\begin{table*}[t]
\centering
\caption{\textbf{Editing results.}
$\Delta$IoU and $\Delta$MSE measure editing precision (spatial and pixel-level change correctness);
Micro/Macro IoU assess final layout quality;
GED evaluates distributional realism; 
Node~F1 and Edge~Overlap assess graph-level consistency.}
\label{tab:editing}
\setlength{\tabcolsep}{6pt}
\begin{tabular}{lccccccccc}
\toprule
\textbf{Method} &
\boldmath$\Delta$\textbf{IoU}&
\boldmath$\Delta$\textbf{MSE}$\downarrow$ &
\textbf{Micro IoU} &
\textbf{Macro IoU} &
\textbf{GED}$\downarrow$ &
\textbf{Node F1} &
\textbf{Edge Ovl.} &
\textbf{Time (s)} \\
\midrule
Before edit                 & --    & --    & 0.880 & 0.821 & 3.06 & 0.934 & 0.740 & -- \\
\midrule
FLUX.1-Kontext-dev   & 0.053 & 0.0162 & 0.289 & 0.185 & 8.91 & 0.765 & 0.222 & $\sim$240 \\
Qwen-Image-Edit-2509 & 0.088 & 0.0074 & 0.567 & 0.429 & 7.96 & 0.915 & 0.426 & $\sim$240 \\
\midrule
HouseMind-E               & \textbf{0.608} & \textbf{0.0019} & \textbf{0.855} & \textbf{0.823} & \textbf{0.467} & \textbf{0.998} & \textbf{0.934} & $\sim$3 \\
HouseMind-O            & 0.598 & 0.0022 & 0.844 & 0.813 & 0.653 & 0.997 & 0.908 & $\sim$3 \\
\bottomrule
\end{tabular}
\end{table*}

\vspace{3pt}
\noindent\textbf{Stage 2. Multimodal pre-training.}
After establishing the shared vocabulary, the model is trained on large-scale paired data comprising textual descriptions, outline tokens, and room tokens.  
Using an autoregressive language-modeling objective, the model learns to predict the next token in mixed sequences of text and spatial tokens.  
This stage enables bidirectional alignment between language and geometry: the model learns to interpret textual spatial relations (e.g., “the kitchen is north of the living room”) and to reconstruct or complete geometric layouts from tokenized representations.  
Through this process, the model acquires a unified understanding of architectural semantics and geometry, serving as the foundation for subsequent instruction tuning and spatial reasoning.

\vspace{3pt}
\noindent\textbf{Stage 3. Instruction tuning (supervised fine-tuning, SFT).}
The final stage performs SFT on curated multimodal instruction data.
It covers three core tasks:
(1) Understanding: interpreting existing layouts to describe room topology and relations;  
(2) Generation: synthesizing plausible layouts from text and outlines;  
(3) Editing: modifying existing floor plans according to natural-language instructions.  
This SFT stage grants the model task awareness, spatial reasoning skills, and controllability, enabling consistent and prompt-driven design interaction.

\vspace{3pt}
\noindent
Each training sample is serialized into an interleaved sequence of text and spatial tokens, allowing the model to learn within a unified autoregressive framework. During inference, the same formulation supports tasks such as text-to-layout generation, layout interpretation, and instruction-based editing within a single unified architecture.

\section{Experiments}
\label{sec:exp}

\subsection{Benchmark Construction and Data Processing}
\label{subsec:benchmark}

We first construct a canonical JSON representation for each floor plan, encoding room type, area, centroid, and pairwise spatial relations. 
Based on these JSONs, Qwen3-30B-A3B~\cite{qwen3} automatically generates two textual descriptions per sample: 
a simple version summarizing the layout and a detailed version including areas, positions, and relations. 
The two versions are mixed to form a linguistically diverse corpus that serves as the pre-training base for \textbf{HouseMind}. 

Built upon this base, we establish the first unified benchmark that jointly evaluates understanding, generation, and editing of architectural floor plans under consistent geometry, text, and evaluation protocols.  
The understanding and generation tasks are directly derived from the mixed corpus, while the editing subset extends it with controlled structural modifications such as adding or removing rooms to assess spatial controllability.

In total, 2,308 samples are reserved as a shared test set for all three tasks, while the remaining 76,122 and 2,308 samples are used for training and validation, respectively.

\subsection{Evaluation Protocols and Metrics}
\label{subsec:metrics}

We adopt a unified evaluation protocol across all tasks, measuring both pixel-level geometry and graph-level spatial consistency. 
Pixel-level metrics (e.g., Micro/Macro~IoU, SSIM~\cite{ssim}, PSNR, FID~\cite{fid}) assess geometric and perceptual fidelity of the generated layouts, 
while structure-level metrics (e.g., Node~F1, Edge~Overlap, GED~\cite{ged}) evaluate topological correctness and relational consistency among rooms. 
This ensures that the reported scores reflect only meaningful and structurally valid predictions.

All evaluations are performed on $256{\times}256$ color-mapped layouts that contain wall boundaries.
We apply a graphics-based post-processing method to normalize all generated images to the same size and append wall boundaries. Results are averaged over the shared test set of 2,308 samples. Since this study aims to develop a lightweight and locally deployable room layout design method, all compared approaches in this section are implemented to run on a single NVIDIA RTX~3090 GPU, and inference time is reported in seconds per sample. Implementation details of other methods are provided in the supplementary materials.

\begin{figure*}[t]
    \centering
    \includegraphics[width=\linewidth]{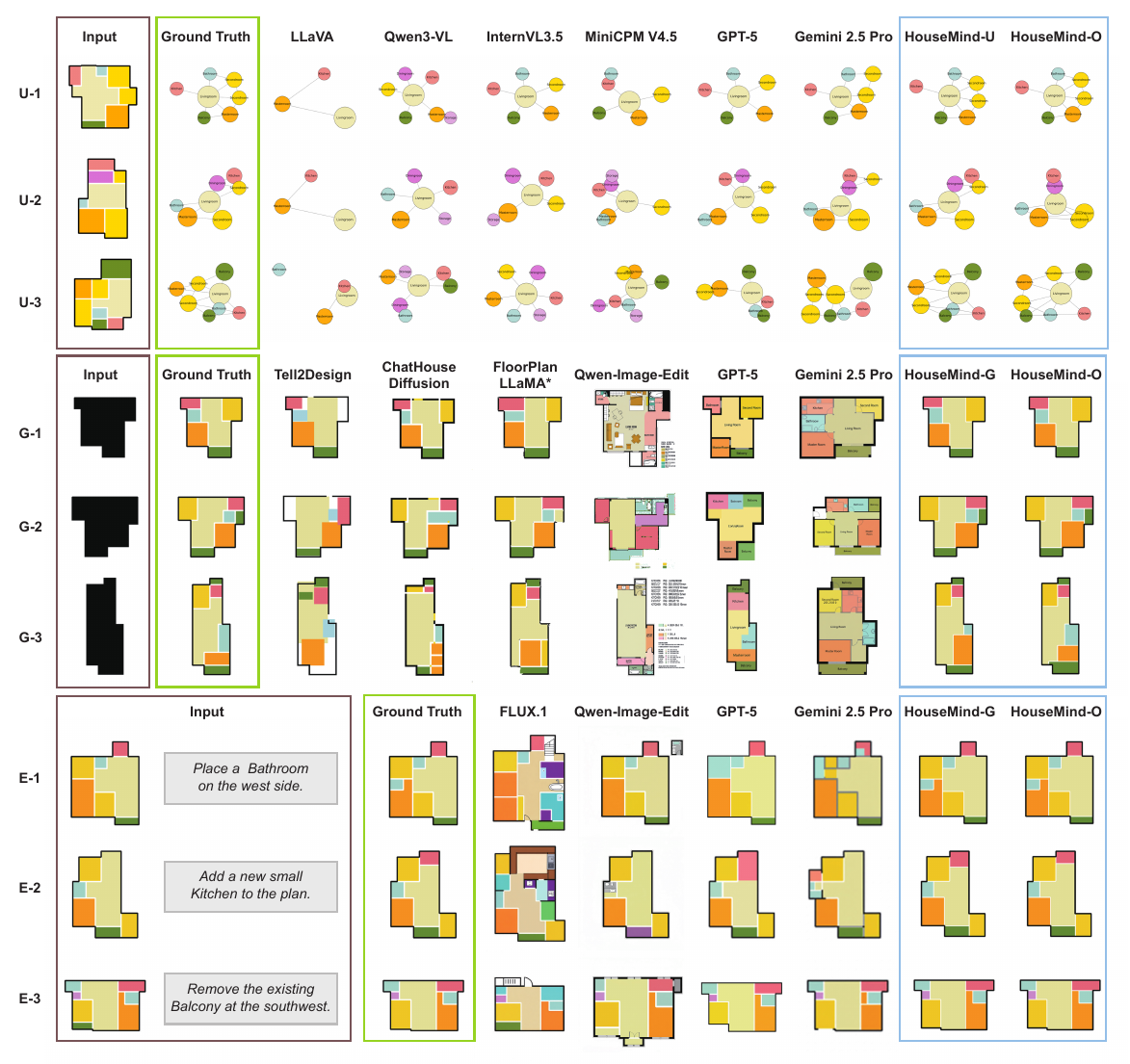}
    \caption{Qualitative comparison results of understanding, generation, and editing.
For understanding tasks (U), \textbf{HouseMind} accurately identifies the number of rooms and their connections.
For generation tasks (G), \textbf{HouseMind} preserves both the room layout and the overall outline consistency; the generation prompts are provided in the supplementary materials.
For editing tasks (E), \textbf{HouseMind} accurately executes the specified modifications when the instructions are explicit.}
    \label{fig:result}
\end{figure*}

\subsection{Quantitative Analysis}
\label{subsec:quantitative}
We train separate models for the three core tasks: \textbf{HouseMind-U} (understanding), \textbf{HouseMind-G} (generation), \textbf{HouseMind-E} (editing), and a unified variant, \textbf{HouseMind-O} (Omni), jointly trained on all tasks for better cross-task generalization.
All models share the same hierarchical tokenization and architectural backbone based on Qwen3-0.6B~\cite{qwen3}, ensuring consistent and efficient multimodal reasoning.

\vspace{3pt}
\noindent\textbf{Understanding.}
Understanding architectural layouts requires not only recognizing room types but also reasoning about their spatial hierarchy and inter-room relationships, aligning visual geometry with textual semantics. 
Current vision–language models (e.g., LLaVA-v1.6-Mistral-7B-HF~\cite{llava}, Qwen3-VL-8B-Instruct~\cite{bai2025qwen3vltechnicalreport}, InternVL3.5-8B~\cite{wang2025internvl3_5}, MiniCPM-V~4.5~\cite{yao2024minicpm}) rely on large-scale multimodal pre-training and show strong visual perception, but they lack structured reasoning and spatial consistency at the room-instance level. To overcome these issues, \textbf{HouseMind} introduces explicit structural and semantic constraints for more accurate and consistent spatial understanding.

As shown in Table~\ref{tab:understanding}, \textbf{HouseMind} achieves superior performance across all metrics, reaching perfect success and matching rates, indicating precise multimodal alignment between geometry and language. 
Compared with vision--language baselines, it improves room localization and adjacency accuracy by more than 40 absolute points and reduces the mean room area error from several square meters to below~0.6 $m^2$. 
Its relational reasoning accuracy of about~0.8 further demonstrates the model's ability to infer complex spatial dependencies beyond direct visual cues. 
Moreover, \textbf{HouseMind-O} performs comparably to its individually trained counterpart, suggesting that the unified multitask architecture maintains both robustness and efficiency.

\vspace{3pt}
\noindent\textbf{Generation.}
Unlike general-purpose diffusion or image-based models (e.g., ChatHouseDiffusion~\cite{qin2024chathousediffusion}, FloorPlanLLaMA~\cite{yinFloorPlanLLaMaAligningArchitects2025}) and open-source multimodal image editing models (e.g., Qwen-Image-Edit-2509~\cite{wu2025qwenimagetechnicalreport}, 
FLUX.1-Kontext-dev~\cite{labs2025flux1kontextflowmatching}), \textbf{HouseMind} maintains geometric validity, semantic consistency, and topological coherence throughout the generation process. For fair comparison, all methods are evaluated under a unified graphical post-processing pipeline that standardizes geometric and scale alignment, as detailed in the supplementary materials.

Notably, FLUX.1-Kontext-dev failed to produce complete layouts owing to the CLIP encoder’s restricted token budget, while Qwen-Image-Edit-2509 was executed under int8 quantization to fit GPU memory constraints.

As summarized in Table~\ref{tab:generation}, \textbf{HouseMind} consistently outperforms existing methods across both pixel-level and graph-level metrics. It achieves Micro/Macro IoU scores of 0.71/0.65, improving IoU by over 10\% compared with ChatHouseDiffusion, while reducing FID from 11.3 to 1.9, demonstrating significant gains in realism and spatial precision. Graph-based indicators further confirm that \textbf{HouseMind} generates layouts with superior room connectivity and adjacency consistency. The unified multimodal variant, \textbf{HouseMind-O}, achieves even more stable results across all metrics, validating the effectiveness of unified pre-training for multimodal spatial reasoning and generation.

\vspace{3pt}
\noindent\textbf{Editing.}  
For the editing task, \textbf{HouseMind} jointly encodes textual editing instructions and the original layout under explicit geometric and semantic constraints, enabling precise additive and subtractive modifications while maintaining global spatial logic. Unlike general-purpose image-editing models, it performs structure-aware control rather than low-level pixel manipulation.

As shown in Table~\ref{tab:editing}, this design yields controllable and spatially consistent modifications with significantly higher editing fidelity compared to the post-edit ground truth. Since some editing instructions do not explicitly specify room positions or sizes, we also evaluate structural-level consistency. The nearly perfect Node~F1 indicates accurate modification of room types, while the substantially lower GED and higher edge overlap relative to the pre-edit layouts demonstrate that \textbf{HouseMind} produces more coherent and semantically consistent spatial relationships after editing. In contrast, general multimodal image editing models fail to achieve comparable results and often degrade the original structural integrity.

\subsection{Qualitative Analysis}
To further evaluate the effectiveness of our approach, we conduct a qualitative comparison using two state-of-the-art multimodal models, GPT-5~\cite{IntroducingGPT52025} and Gemini~2.5~Pro~\cite{comanici2025gemini25}, as shown in Fig.~\ref{fig:result}.

\vspace{3pt}
\noindent\textbf{Understanding.}
LLaVA-v1.6-Mistral-7B-HF tends to output a fixed pattern and fails to produce complete layouts. Qwen3-VL-8B-Instruct generates almost all room types regardless of the actual input, resulting in a mismatch with the ground truth. InternVL3.5-8B and MiniCPM-V~4.5 achieve more accurate room prediction, yet they still suffer from noticeable errors in room size estimation and topological relationships.
GPT-5 achieves a noticeable improvement, while Gemini~2.5~Pro produces relatively coherent bubble-like representations but still contains room-type errors. 
In contrast, \textbf{HouseMind} captures room types, sizes, and topological relations with high fidelity, exhibiting only minor local inaccuracies.

\vspace{3pt}
\noindent\textbf{Generation.}
Tell2Design is trained on manually annotated and rule-based datasets, which limits its generalization to out-of-distribution layouts. 
ChatHouseDiffusion leverages room outlines effectively and performs well on simple layouts but struggles with complex spatial configurations. 
FloorPlanLLaMA employs a VQ-VAE to encode the entire floor plan; while it preserves the approximate room count, it often fails to maintain boundary consistency and produces noticeable noise (the results shown are after post-processing). 
Qwen-Image-Edit generates outputs inconsistent with the given prompts. 
GPT-5 and Gemini~2.5~Pro produce generally reasonable room layouts but still fail to meet specific design constraints. 
By contrast, \textbf{HouseMind} generates results that align well with the textual descriptions and accurately conform to the given building outlines. 

\vspace{3pt}
\noindent\textbf{Editing.}
The editing task is relatively simple; however, both FLUX.1-Kontext-dev and Qwen-Image-Edit-2509 tend to introduce irrelevant elements during generation. 
GPT-5 performs better and generally fulfills the editing instructions but slightly alters the existing layout, while Gemini~2.5~Pro performs somewhat worse. 
In contrast, \textbf{HouseMind} achieves precise, localized modifications without affecting unrelated regions, demonstrating fine-grained spatial control and superior structural consistency.

\subsection{Ablations}

All ablation studies are conducted on the validation split to avoid biasing the held-out test set used for Tables~\ref{tab:understanding}–\ref{tab:editing}.
Validation cross-entropy (Eval Loss) is adopted as a concise indicator of cross-modal alignment, where lower values indicate stronger correspondence between text and spatial representations.

As illustrated in Sec.~\ref{subsec:housemind}, \textbf{HouseMind} follows a three-stage training pipeline.
To evaluate the contribution of each stage, we compare four variants: removing both Stage~1 and~2 (w/o Stage~1\&2), removing Stage~1 only (w/o Stage~1), removing Stage~2 only (w/o Stage~2), and the full model (Full).

\begin{table}[ht]
\centering
\caption{Loss under different training-stage configurations.}
\label{tab:ablation_loss}
\setlength{\tabcolsep}{10pt}
\begin{tabular}{lcc}
\toprule
\textbf{Model} & \textbf{Train Loss} $\downarrow$ & \textbf{Eval Loss} $\downarrow$ \\
\midrule
w/o Stage~1 \& 2 & 0.0729 & 0.0836 \\
w/o Stage~1      & 0.0659 & 0.0840 \\
w/o Stage~2      & 0.0712 & 0.0831 \\
\textbf{Full} & \textbf{0.0644} & \textbf{0.0830} \\
\bottomrule
\end{tabular}
\end{table}

Removing either Stage~1 or Stage~2 increases the evaluation loss, revealing weaker multimodal alignment. 
Without Stage~1, the model fails to ground spatial tokens within a stable embedding space, causing optimization instability. 
Without Stage~2, the backbone lacks higher-level text–layout correspondence even when embeddings are initialized. 
The complete three-stage pipeline achieves the lowest loss, demonstrating that Stage~1 ensures consistent token-level initialization, while Stage~2 refines global spatial reasoning. 
Together, this progressive alignment process underscores the necessity of coupling stable spatial grounding with high-level semantic fusion for robust multimodal understanding. It provides the foundation that enables \textbf{HouseMind} to generalize effectively across understanding, generation, and editing tasks.

\section{Discussion}
\label{sec:dis}
\noindent\textbf{Limitations.}
Despite promising performance, several limitations remain. 
(1) The current editing module focuses primarily on simple operations such as room addition and deletion, without supporting complex topological transformations. 
(2) Functional components like doors, windows, and furniture are not yet modeled, limiting the model’s applicability to detailed interior design. 
(3) The system’s behavior is not fully aligned with human design preferences and aesthetic constraints, leaving a gap between AI-generated layouts and professional design standards.

\vspace{3pt}
\noindent\textbf{Future Directions.}
Future work will address these issues from multiple perspectives. 
(1) Expand the dataset~\cite{vanengelenburgMSDBenchmarkDataset2025} and instruction set to include a broader variety of editing tasks, improving generalization across unseen layout patterns. 
(2) Integrate architecture-structure co-design~\cite{lengArchiDiffusionNovelDiffusion2024} into the pipeline, enabling full-process generative design that spans architectural layout and structural configuration. 
(3) Incorporate performance evaluation metrics~\cite{meselhyReviewArtificialIntelligence2025} and human-centered preference alignment~\cite{yinFloorPlanLLaMaAligningArchitects2025}, allowing the model to generate layouts that not only satisfy spatial logic but also meet safety, comfort, and sustainability criteria.

\section{Conclusion}
\label{sec:conc}
\textbf{HouseMind} is a unified and lightweight framework that enables MLLMs to understand, generate, and edit architectural floor plans through hierarchical tokenization. 
By bridging geometric structures with linguistic reasoning, it achieves coherent, controllable, and interpretable spatial design. 
Its room-by-room reasoning paradigm aligns with how architects iteratively conceive and refine functional spaces in practice. 
Extensive experiments across understanding, generation, and editing tasks demonstrate consistent gains in accuracy, efficiency, and semantic fidelity. 
These results establish tokenization as a key mechanism linking large language models with spatial design intelligence, 
marking a crucial step toward human-aligned and performance-aware architecture--structure co-design.

\section*{Acknowledgements}
This work was supported by the Beijing Municipal Natural Science Foundation (8252008), the Tsinghua University Initiative Scientific Research Program (2025Z03KYY001), and the National Natural Science Foundation of China (525B2130).
{
    \small
    \bibliographystyle{ieeenat_fullname}
    \bibliography{main}
}

\twocolumn[\newpage]
\appendix
\renewcommand{\thefigure}{A\arabic{figure}}
\renewcommand{\thetable}{A\arabic{table}}
\renewcommand{\theequation}{A.\arabic{equation}}
\numberwithin{figure}{section}
\numberwithin{table}{section}
\numberwithin{equation}{section}
\setcounter{figure}{0}
\setcounter{table}{0}
\section{Implementation Details}
\subsection{Room-Instance Tokenization}

\begin{figure}[ht]
\centering
\includegraphics[width=\linewidth]{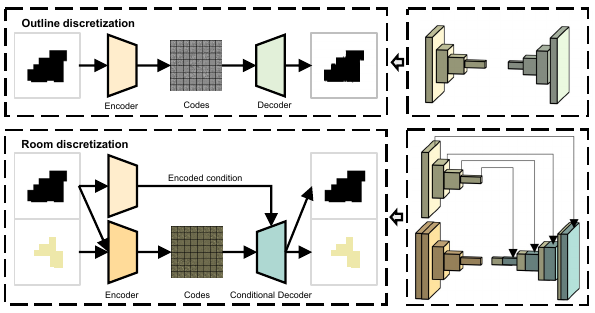}
\caption{
\textbf{VQ-VAE tokenization framework.}
The outline branch encodes the global boundary, 
while the conditional room branch encodes each room with outline context to capture spatial relations.
}
\label{fig:vqvae}
\end{figure}

We employ two lightweight VQ-VAE branches (Fig.~\ref{fig:vqvae}) to discretize floor plan geometry into spatial tokens: 
an \textbf{outline branch} that encodes the global building boundary, 
and a \textbf{conditional room branch} that encodes each room conditioned on its corresponding outline. 
Specifically, the room encoder concatenates the outline mask as an additional input channel to preserve adjacency and boundary consistency. 
Together, these branches produce a hierarchical spatial vocabulary serving as the foundation for multimodal alignment and instruction tuning.

We study reconstruction fidelity with respect to token granularity and codebook capacity. 
As shown in Tables~\ref{tab:vqvae-outline} and~\ref{tab:vqvae-room}, 
token count strongly influences downstream sequence length and contextual reasoning ability: 
$2\times2$ tokens underfit geometry, while $16\times16$ tokens generate unnecessarily long sequences. 
Hence, we focus on $4\times4$ and $8\times8$ configurations for practical balance.  

\begin{table}[ht]
\centering
\caption{Outline branch: PSNR (dB) and SSIM under different token grids and codebook sizes.}
\label{tab:vqvae-outline}
\vspace{2mm}
\begin{tabular}{cccc}
\toprule
$\mathbf{N_{\text{\textbf{tokens}}}}$ & \textbf{Codebook} & \textbf{PSNR} & \textbf{SSIM} \\
\midrule
$4\times4$ & 256  & 20.003 & 0.911 \\
$4\times4$ & 512  & 22.188 & 0.946 \\
$4\times4$ & 1024 & 23.202 & 0.957 \\
$8\times8$ & 256  & 35.898 & 0.998 \\
$8\times8$ & 512  & 36.657 & 0.998 \\
$8\times8$ & 1024 & 37.053 & 0.998 \\
\bottomrule
\end{tabular}
\end{table}

\begin{table}[ht]
\centering
\caption{Room branch: PSNR (dB) and SSIM under different token grids and codebook sizes.}
\label{tab:vqvae-room}
\vspace{2mm}
\begin{tabular}{cccc}
\toprule
$\mathbf{N_{\text{\textbf{tokens}}}}$ & \textbf{Codebook} & \textbf{PSNR} & \textbf{SSIM} \\
\midrule
$4\times4$ & 256  & 31.896 & 0.994 \\
$4\times4$ & 512  & 33.106 & 0.995 \\
$4\times4$ & 1024 & 34.506 & 0.997 \\
$8\times8$ & 256  & 44.624 & 1.000 \\
$8\times8$ & 512  & 48.467 & 1.000 \\
$8\times8$ & 1024 & 51.769 & 1.000 \\
\bottomrule
\end{tabular}
\end{table}

Since each outline corresponds to multiple room instances, the room-level data volume is substantially larger, 
necessitating a smaller learning rate and fewer epochs for stable convergence. 
Considering the trade-off between fidelity and efficiency, $8\times8$ tokens deliver clearly superior reconstruction to $4\times4$ 
while avoiding the context-length overhead of $16\times16$. 
Although increasing the codebook size yields modest gains, 
a compact codebook of 256 effectively balances information density and vocabulary size.  

Concretely, both branches adopt a latent dimension of 256 with three downsampling layers. 
The outline VQ-VAE is trained for 50 epochs with batch size 256 and learning rate $3\times10^{-4}$, 
while the conditional room VQ-VAE is trained for 30 epochs with batch size 256 and learning rate $1\times10^{-4}$.

A potential concern is whether the VQ-VAE discretization introduces an information bottleneck that limits downstream generation quality. 
To assess this, we first examine reconstruction fidelity. 
As the reconstruction PSNR is controlled around 40 dB, it is already close to a near-lossless level for layout geometry. Qualitative comparisons in Fig.~\ref{fig:vqvae} show that reconstructed layouts are almost indistinguishable from the original inputs, indicating minimal information loss during tokenization.

We further conduct a codebook size ablation study to evaluate whether generation performance is sensitive to token vocabulary capacity. 
Results are summarized in Table~\ref{tab:codebook}. 
Across codebook sizes ranging from 256 to 1024, 
generation metrics (Macro IoU, FID, GED, Node F1, and Edge Overlap) remain largely stable. 
The performance variation is marginal, suggesting that the discretized latent space preserves sufficient structural information for downstream reasoning.

These results indicate that the VQ-VAE module does not constitute a performance bottleneck in our pipeline.

\begin{table}[htbp]
\centering
\caption{Codebook size ablation study}
\label{tab:codebook}
\setlength{\tabcolsep}{5pt}
\begin{tabular}{cccccccccc}
\toprule
\textbf{Size} &
\textbf{Ma. IoU} & 
\textbf{FID}$\downarrow$ & 
\textbf{GED}$\downarrow$ & 
\textbf{Node F1} & 
\textbf{Edge Ovl.}\\
\midrule
256             & 0.654 & \textbf{1.89} & 1.03 & 0.994 & 0.880 \\
512             &\textbf{0.657} & 1.95 & 1.04 & \textbf{0.995} & 0.877 \\
1024            &\textbf{0.657} & 1.97 & \textbf{1.00} & 0.994 & \textbf{0.881} \\
\bottomrule
\end{tabular}
\end{table}

\subsection{Multimodal Alignment and Instruction Tuning}
LLM training is conducted on Ubuntu~22.04.5~LTS with an AMD~EPYC~7K62~48-Core~CPU (96~threads), 256~GB~RAM, and an NVIDIA~GeForce~RTX~5090~GPU (32~GB~VRAM). 
All experiments are implemented using the LLaMA-Factory framework with the Qwen3-0.6B backbone and FlashAttention-2 acceleration. 
To integrate spatial information, we extend the original tokenizer by manually appending a set of discrete outline and room tokens derived from the VQ-VAE codebooks, enabling the LLM to process geometric and semantic content within a unified vocabulary. 
Each multimodal sample contains up to 2048~tokens, including spatial tokens and textual instructions. 
The model is trained in three sequential stages following the pipeline described in Sec. 4.2:
embedding initialization, multimodal pre-training, and instruction tuning. 
A cosine learning-rate schedule with 10\% warm-up is adopted throughout. 
Early stages employ a small effective batch size (4~with gradient accumulation) to stabilize optimization under long-sequence multimodal inputs, while Stage~3 uses a larger batch size (16) for instruction-level generalization. 
Learning rates are set to $1\times10^{-5}$ for Stages~1--2 and $2\times10^{-5}$ for Stage~3, with 1,~2,~and~3~epochs respectively.

\subsection{Baseline Methods}
To ensure a fair comparison, all baseline methods are reproduced or standardized under a unified training and inference pipeline. 
All models are evaluated at $256\times256$ resolution with wall-boundary restoration, and inference is conducted on a single RTX~3090 GPU (an Intel~Xeon~E5-2682~v4~CPU and 32~GB RAM).

\vspace{3pt}
\noindent\textbf{Understanding.}
For multimodal reasoning baselines, including LLaVA-v1.6-Mistral-7B-HF, Qwen3-VL-8B-Instruct, InternVL3.5-8B, and MiniCPM-V~4.5, we use their official checkpoints and serve them through vLLM for efficient inference. 
Each model receives the color-mapped layout and text instruction, producing structured descriptions for evaluating room-type classification, localization accuracy, and relational reasoning.  
The prompt template used for understanding is as follows:

\begin{tcolorbox}[promptstyle, title={Prompt Template for Understanding:}]
\footnotesize\ttfamily
\obeylines
You are an expert in architectural layout understanding.
The input is a color-coded floor plan image with an overall size of 18m $\times$ 18m.
Each color approximately represents a functional area, and the full layout fits within this boundary.

Color-to-room mapping:
Light khaki or pale yellow $\rightarrow$ Livingroom (large open space)
Orange $\rightarrow$ Masterroom (private sleeping area)
Light salmon or red $\rightarrow$ Kitchen (food preparation area)
Light cyan $\rightarrow$ Bathroom (toilet or shower area)
Violet $\rightarrow$ Diningroom (area for meals)
Plum $\rightarrow$ Storage (small enclosed area)
Bright yellow $\rightarrow$ Commonroom (Secondroom / Studyroom / Childroom / Guestroom)
Olive green $\rightarrow$ Balcony (semi-outdoor area)
Black $\rightarrow$ Exterior wall or building boundary
White $\rightarrow$ Front door, main entrance, interior walls, interior doors, and external background

Example:
Below is an example floor plan analysis.

Input image: [example\_floorplan.png]
Output JSON:
\{
  'rooms': [
    \{'idx': 0, 'type': 'LivingRoom', 'area': 33, 'width': 6, 'height': 9, 'position': 'east'\},
    \{'idx': 1, 'type': 'SecondRoom', 'area': 12, 'width': 3, 'height': 4, 'position': 'northwest'\},
    \{'idx': 2, 'type': 'MasterRoom', 'area': 12, 'width': 3, 'height': 5, 'position': 'southwest'\},
    \{'idx': 3, 'type': 'StudyRoom', 'area': 11, 'width': 3, 'height': 4, 'position': 'north'\},
    \{'idx': 4, 'type': 'Bathroom', 'area': 4, 'width': 2, 'height': 2, 'position': 'west'\},
    \{'idx': 5, 'type': 'Kitchen', 'area': 4, 'width': 2, 'height': 2, 'position': 'northeast'\}
  ],
  'edges': [
    \{'room1': 5, 'room2': 3, 'relation': 'right-of', 'text': 'Kitchen is right-of StudyRoom'\},
    \{'room1': 5, 'room2': 0, 'relation': 'above', 'text': 'Kitchen is above LivingRoom'\},
    \{'room1': 1, 'room2': 3, 'relation': 'left-of', 'text': 'SecondRoom is left-of StudyRoom'\},
    \{'room1': 1, 'room2': 4, 'relation': 'above', 'text': 'SecondRoom is above Bathroom'\},
    \{'room1': 4, 'room2': 2, 'relation': 'above', 'text': 'Bathroom is above MasterRoom'\}
  ],
  'description': 'The floor plan centers around the spacious Living Room, with surrounding rooms arranged by clear spatial logic.'
\}

Now follow this example format for the next images.

Instruction: Provide both room (node) attributes and spatial (edge) relations in JSON format.

Your task:
- Identify all rooms (nodes) within the 18m $\times$ 18m plan.
- For each room, estimate attributes:
  - idx (int): unique ID
  - type (str): functional category (e.g., LivingRoom, Kitchen)
  - area (float): area in square meters
  - width, height (float): dimensions in meters
  - position (str): coarse location in the 18m $\times$ 18m layout (e.g., north, center, southeast)
- Infer spatial relations (edges) between rooms using ONLY the following edge types:
  ['left-above', 'left-below', 'left-of', 'above', 'inside', 'surrounding', 'below', 'right-of', 'right-above', 'right-below']
- Each edge must:
  - reference valid room indices (room1, room2)
  - use exactly one relation from the list above as "relation"
  - include a short human-readable "text"

Semantics (guidance):
- "left-of"/"right-of"/"above"/"below": strict axis-aligned relations.
- "left-above", "left-below", "right-above", "right-below": diagonal/oblique relations combining horizontal and vertical.
- "inside": room1 is fully inside room2 (room2 acts as container).
- "surrounding": room1 encloses or wraps around room2 (the inverse of "inside").

Output format (must be valid JSON; use double quotes):
\{
  "rooms": [\{"idx": 0, "type": "RoomType", "area": 0.0, "width": 0.0, "height": 0.0, "position": "center"\}],
  "edges": [\{"room1": 0, "room2": 1, "relation": "left-of", "text": "Room0 is left-of Room1"\}],
  "description": "One-sentence summary of the layout."
\}

Requirements:
- Use only the allowed edge types listed above.
- All dimensions/positions are interpreted relative to the 18m $\times$ 18m plan.
- Output a single valid JSON object and nothing else.

\end{tcolorbox}

\vspace{3pt}
\noindent\textbf{Generation.}
Tell2Design and ChatHouseDiffusion are reproduced using their released implementations. 
Tell2Design performs direct text-to-layout generation, while ChatHouseDiffusion employs diffusion sampling with language conditioning. 
Since the official FloorPlan-LLaMA is not publicly available, we reproduce it using a VQ-VAE tokenizer and Qwen backbone. 
Floor plans are converted to grayscale and tokenized with $16\times16$ latent grids, codebook size~1024, latent dimension~256.
The tokenizer reaches approximately 38.1~dB PSNR and 0.99~SSIM, confirming high geometric fidelity before language alignment. 
Qwen-Image-Edit-2509 is also included as a generation baseline.  
The prompt template used for generation is:

\begin{tcolorbox}[promptstyle, title={Prompt Template for Generation:}]
\footnotesize\ttfamily
\obeylines

You are an expert in architectural layout generation.
The input image is a black building outline.
Please generate a complete flat color-blocked floor plan \textbf{within the provided outline}.

- Keep the wall boundaries fixed and fill enclosed regions with appropriate colors.
- Each room or functional area should be represented by a solid color block according to the color legend.
- The generated layout must align with the spatial description below.
- Avoid adding any text, labels, furniture, shadows, or perspective effects.

Spatial Description:
[SPATIAL\_DESCRIPTION]

Color Legend:
Light khaki (RGB 238 232 170) $\rightarrow$ Livingroom (large open space), Entrance, Wall-in
Orange (RGB 255 165 0) $\rightarrow$ Masterroom (private sleeping area)
Light salmon (RGB 240 128 128) $\rightarrow$ Kitchen (food preparation area)
Light cyan (RGB 173 216 210) $\rightarrow$ Bathroom (toilet or shower area)
Olive green (RGB 107 142 35) $\rightarrow$ Balcony (semi-outdoor area)
Violet (RGB 218 112 214) $\rightarrow$ Diningroom (area for meals)
Plum (RGB 221 160 221) $\rightarrow$ Storage (small enclosed area)
Bright yellow (RGB 255 215 0) $\rightarrow$ Commonroom / Secondroom / Studyroom / Childroom / Guestroom
Black (RGB 0 0 0) $\rightarrow$ Exterior wall or building boundary
White (RGB 255 255 255) $\rightarrow$ Front door / main entrance / interior walls / interior doors / external background

Output Goal:
Produce a visually clean and semantically accurate colored floor plan image that fits exactly within the input outline.
\end{tcolorbox}

\vspace{3pt}
\noindent\textbf{Editing.}
For structure-aware layout editing, we evaluate FLUX.1-Kontext-dev and Qwen-Image-Edit-2509 using the original layout and a textual edit command. The prompt template for editing is:

\begin{tcolorbox}[promptstyle, title={Prompt Template for Editing:}]
\footnotesize\ttfamily
\obeylines
You are an expert in architectural floor plan editing.
The input image is a \textbf{color-coded floor plan} representing functional areas with flat color blocks.

Your task is to \textbf{edit the given layout} based on the following instruction,
while maintaining architectural coherence and color consistency.

Editing Guidelines:
- Follow the edit instruction carefully (e.g., add, remove, move, merge, resize rooms).
- Preserve unrelated regions and overall wall boundaries.
- Use the same color scheme for all room types as specified in the legend.
- Maintain clean edges and closed regions without overlapping or blending.
- Do not add any text, furniture, or decorations.
- Ensure that all colors remain consistent with the color legend below.

Edit Instruction:
[EDIT\_INSTRUCTION]

Color Legend:
Light khaki (RGB 238 232 170) $\rightarrow$ Livingroom (large open space), Entrance, Wall-in
Orange (RGB 255 165 0) $\rightarrow$ Masterroom (private sleeping area)
Light salmon (RGB 240 128 128) $\rightarrow$ Kitchen (food preparation area)
Light cyan (RGB 173 216 210) $\rightarrow$ Bathroom (toilet or shower area)
Olive green (RGB 107 142 35) $\rightarrow$ Balcony (semi-outdoor area)
Violet (RGB 218 112 214) $\rightarrow$ Diningroom (area for meals)
Plum (RGB 221 160 221) $\rightarrow$ Storage (small enclosed area)
Bright yellow (RGB 255 215 0) $\rightarrow$ Commonroom / Secondroom / Studyroom / Childroom / Guestroom
Black (RGB 0 0 0) $\rightarrow$ Exterior wall or building boundary
White (RGB 255 255 255) $\rightarrow$ Front door / main entrance / interior walls / interior doors / external background

Output Goal:
Return an updated floor plan image with \textbf{only the described edits applied}, preserving the rest of the layout unchanged.
\end{tcolorbox}

\section{Evaluation Metrics}
\label{app:metrics}

We evaluate model performance for three tasks using unified geometric and structural metrics.
Each floor plan is a $256{\times}256$ color-coded layout corresponding to an $18{\times}18$\,m area.

\vspace{3pt}
\subsection{Understanding Metrics}

\noindent\textbf{Success Rate (Success).}
\begin{equation}
\mathrm{Success} = \frac{N_{\mathrm{success}}}{N_{\mathrm{total}}},
\end{equation}
where $N_{\mathrm{success}}$ and $N_{\mathrm{total}}$ are the numbers of successful and total samples.

\vspace{3pt}\noindent
\textbf{Room Match Rate (RMR).}
\begin{equation}
\mathrm{RMR} = \frac{1}{N}\sum_{i=1}^{N} I(t_i = \hat{t}_i),
\end{equation}
where $N$ is the number of rooms, 
$t_i$ and $\hat{t}_i$ are the ground-truth and predicted room types,
and $I(\cdot)$ equals 1 if the condition is true and 0 otherwise.

\vspace{3pt}\noindent
\textbf{Room Location Accuracy (LocAcc).}
\begin{equation}
\mathrm{LocAcc} = \frac{1}{N}\sum_{i=1}^{N} I(p_i = \hat{p}_i),
\end{equation}
where $p_i$ and $\hat{p}_i$ denote the coarse positions of room $i$ 
(e.g., north, south, east, west, or center).

\vspace{3pt}\noindent
\textbf{Room Area Difference (AreaDiff).}
\begin{equation}
\mathrm{AreaDiff} = \frac{1}{N}\sum_{i=1}^{N} |A_i - \hat{A}_i|,
\end{equation}
where $A_i$ and $\hat{A}_i$ are the ground-truth and predicted areas (in m$^2$).

\vspace{3pt}\noindent
\textbf{Room Adjacency Accuracy (AdjAcc).}
\begin{equation}
\mathrm{AdjAcc} = \frac{|E \cap \hat{E}|}{|E \cup \hat{E}|}
\end{equation}
where $E$ and $\hat{E}$ denote the ground-truth and predicted adjacency sets.

\vspace{3pt}\noindent
\textbf{Spatial Relation Accuracy (RelAcc).}
\begin{equation}
\mathrm{RelAcc} = \frac{1}{|R|}\sum_{(i,j)\in R} I(r_{ij} = \hat{r}_{ij}),
\end{equation}
where $R$ is the set of all adjacent ordered pairs with directional relations (e.g., left-of, above, below),
and $r_{ij}$ and $\hat{r}_{ij}$ represent their ground-truth and predicted relation types.

\vspace{3pt}
\subsection{Generation Metrics}

\noindent\textbf{Micro IoU.}
\begin{equation}
\mathrm{IoU}_{\mathrm{micro}} =
\frac{\sum_{c=1}^{C} |M_c^{\mathrm{pred}} \cap M_c^{\mathrm{gt}}|}
{\sum_{c=1}^{C} |M_c^{\mathrm{pred}} \cup M_c^{\mathrm{gt}}|},
\end{equation}
where $M_c^{\mathrm{pred}}$ and $M_c^{\mathrm{gt}}$ are binary masks for category $c$.

\vspace{3pt}\noindent
\textbf{Macro IoU.}
\begin{equation}
\mathrm{IoU}_{\mathrm{macro}} =
\frac{1}{C}\sum_{c=1}^{C}
\frac{|M_c^{\mathrm{pred}} \cap M_c^{\mathrm{gt}}|}
{|M_c^{\mathrm{pred}} \cup M_c^{\mathrm{gt}}|}.
\end{equation}
Macro-IoU balances room frequencies by averaging across categories.

\vspace{3pt}\noindent
\textbf{Structural Similarity (SSIM).}
\begin{equation}
\mathrm{SSIM} =
\frac{(2\mu_x\mu_y + C_1)(2\sigma_{xy} + C_2)}
{(\mu_x^2 + \mu_y^2 + C_1)(\sigma_x^2 + \sigma_y^2 + C_2)},
\end{equation}
where $\mu_x,\mu_y$ are mean intensities, 
$\sigma_x,\sigma_y$ variances, and $\sigma_{xy}$ covariance.
$C_1,C_2$ are small constants for stability.

\vspace{3pt}\noindent
\textbf{Peak Signal-to-Noise Ratio (PSNR).}
\begin{equation}
\mathrm{PSNR} = 10\log_{10}\!\left(\frac{L^2}{\mathrm{MSE}}\right),
\end{equation}
where $L$ is the maximum intensity (255) and $\mathrm{MSE}$ is the mean squared error.

\vspace{3pt}\noindent
\textbf{Fréchet Inception Distance (FID).}
\begin{equation}
\mathrm{FID} = \|\mu_r-\mu_g\|_2^2 +
\mathrm{Tr}(\Sigma_r+\Sigma_g-2(\Sigma_r\Sigma_g)^{1/2}),
\end{equation}
where $(\mu_r,\Sigma_r)$ and $(\mu_g,\Sigma_g)$ are the feature means and covariances
of real and generated layouts.

\vspace{3pt}\noindent
\textbf{Graph Edit Distance (GED).}
\begin{equation}
\mathrm{GED}(G_r,G_g) =
\min_{\pi} \sum_{o \in \pi} c(o),
\end{equation}
where $G_r$ and $G_g$ are room–adjacency graphs,
$\pi$ is a sequence of edit operations (node/edge insertion, deletion, or substitution),
and $c(o)$ is the edit cost.
We use the \texttt{graph\_edit\_distance()} function in NetworkX.

\vspace{3pt}\noindent
\textbf{Node~F1.}
\begin{equation}
P = \frac{|T_{\mathrm{pred}} \cap T_{\mathrm{gt}}|}{|T_{\mathrm{pred}}|}, 
\quad
R = \frac{|T_{\mathrm{pred}} \cap T_{\mathrm{gt}}|}{|T_{\mathrm{gt}}|},
\end{equation}
\begin{equation}
\mathrm{Node~F1} = \frac{2PR}{P + R},
\end{equation}
where $T_{\mathrm{pred}}$ and $T_{\mathrm{gt}}$ are the sets of predicted and ground-truth room types,
and $P$, $R$ are precision and recall.

\vspace{3pt}\noindent
\textbf{Edge~Overlap.}
\begin{equation}
\mathrm{Edge~Overlap} = 
\frac{|E_{\mathrm{pred}} \cap E_{\mathrm{gt}}|}
{|E_{\mathrm{pred}} \cup E_{\mathrm{gt}}|},
\end{equation}
which measures the proportion of correctly predicted adjacency relations,
reflecting structural consistency.

\vspace{3pt}
\subsection{Editing Metrics}

\noindent\textbf{Edit-Detection IoU ($\Delta$IoU).}
\begin{equation}
M^{\Delta}_{\mathrm{gt}} = T(M^{\mathrm{gt}}_{\mathrm{after}} - M_{\mathrm{before}}), \quad
M^{\Delta}_{\mathrm{pred}} = T(M^{\mathrm{pred}}_{\mathrm{after}} - M_{\mathrm{before}}),
\end{equation}
\begin{equation}
\Delta\mathrm{IoU} =
\frac{|M^{\Delta}_{\mathrm{gt}} \cap M^{\Delta}_{\mathrm{pred}}|}
{|M^{\Delta}_{\mathrm{gt}} \cup M^{\Delta}_{\mathrm{pred}}|},
\end{equation}
where $T(\cdot)$ denotes thresholding of the difference map to obtain the changed region. During computation, pixels corresponding to wall-line areas are excluded to avoid non-editable regions affecting the metric.

\vspace{3pt}\noindent
\textbf{Diff-MSE ($\Delta$MSE).}
\begin{equation}
\Delta\mathrm{MSE} =
\frac{1}{N}\sum_{p=1}^{N}
\big(M^{\Delta}_{\mathrm{gt}}(p) - M^{\Delta}_{\mathrm{pred}}(p)\big)^2,
\end{equation}
where $p$ indexes pixels and $N$ is the total number of pixels.

\vspace{3pt}\noindent
\textbf{Structural Consistency.}
GED, Node~F1, and Edge~Overlap are reused to measure whether
the edited layout maintains correct topology.

\vspace{3pt}
\subsection{Metric Significance}

Structural metrics (GED, Node~F1, Edge~Overlap) and class-balanced Macro-IoU
capture spatial topology and functional balance.
Perceptual metrics (SSIM, PSNR, FID) assess visual realism,
while edit-specific metrics ($\Delta$IoU, $\Delta$MSE) evaluate
the controllability and precision of modifications.
For floor plan tasks, structural metrics provide more meaningful insights into spatial reasoning and design consistency,
while pixel-level metrics serve primarily as complementary references.

\section{Post-processing and Effect Analysis}
\label{app:postprocess}

\subsection{Post-processing Pipeline}

To ensure clean topology, structural consistency, and standardized color encoding,
all generated floor plans are refined through a unified three-stage pipeline
implemented with PyTorch, OpenCV, and NumPy.
This end-to-end process converts discrete token predictions into
visually coherent, evaluation-ready layouts.

\vspace{3pt}\noindent
\textbf{Floor Plan Rendering.}
During the generation stage, model predictions in tokenized form are converted into complete color layouts through a rendering routine implemented in PyTorch.
Conditioned on the building outline, each room is decoded into a binary occupancy map and overlaid in descending order of area to form the full layout.
Each functional region is assigned a base color from the predefined legend, with slight random jitter to enhance visual distinction.
The resulting layout contains black outer walls, white inner separators, and solid-colored functional areas, serving as the standardized input for the subsequent structural correction and standardization steps.

\vspace{3pt}\noindent
\textbf{Structural Correction.}
Ambiguous gray zones and small irregularities are first removed by replacing undefined pixels with the dominant color of neighboring regions.
Morphological open–close operations are then applied within each connected component to smooth protruding edges while preserving room boundaries and overall spatial integrity.

\vspace{3pt}\noindent
\textbf{Standardization and Contour Refinement.}
After correction, each pixel is reassigned to the nearest entry in the predefined color map to ensure consistent semantic encoding across samples.
Contours are redrawn with black outer walls and white inner separators to enhance visual clarity,
and all layouts are resized to $256\times256$ pixels using nearest-neighbor interpolation to maintain discrete color values.

\begin{figure}[ht]
\centering
\includegraphics[width=0.7\linewidth]{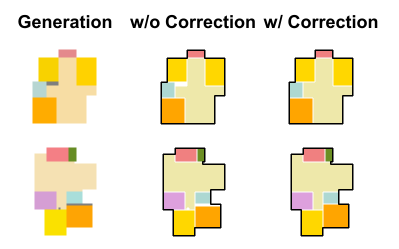}
\caption{Post-processing examples.}
\label{fig:postproc}
\end{figure}

This three-stage pipeline is uniformly applied across all generation and editing experiments for every compared method, ensuring consistent color alignment, structural fidelity, and fair quantitative evaluation under a unified protocol, as illustrated in Fig.~\ref{fig:postproc}.

\subsection{Effect of Structural Correction}

To evaluate the impact of the structural correction step,
we compare generation metrics before and after applying it,
while keeping the standardization process fixed for both settings.
As shown in Table~\ref{tab:postprocess},
structural correction further improves geometric consistency
and perceptual clarity by smoothing irregular edges and resolving
ambiguous regions near boundaries.

\begin{table}[ht]
\centering
\caption{
Impact of Structural Correction in \textbf{HouseMind-O}.
}
\vspace{3pt}
\begin{tabular}{lcc}
\toprule
\textbf{Metric} & \textbf{w/o Correction} & \textbf{w/ Correction} \\
\midrule
Micro IoU    & 0.710 & 0.710 \\
Macro IoU   & 0.654 & 0.654 \\
SSIM         & 0.887 & 0.887 \\
PSNR        & 16.0  & 16.1  \\
FID$\downarrow$        & 1.87  & 1.89  \\
GED$\downarrow$        & 1.10  & 1.03  \\
Node F1      & 0.994 & 0.994 \\
Edge Overlap & 0.873 & 0.880 \\
\bottomrule
\end{tabular}
\label{tab:postprocess}
\end{table}

The corrected results exhibit more coherent room boundaries
and reduced local noise, as reflected by lower GED and higher Edge~Overlap.

\section{Result Analysis and Discussion}

\vspace{3pt}
\subsection{Stability Analysis}

\noindent
To evaluate the robustness and consistency of different models,
we visualize the distribution of evaluation metrics across the entire test set using \textbf{HouseMind-O}.
\begin{figure*}[t]
    \centering
    \includegraphics[width=\linewidth]{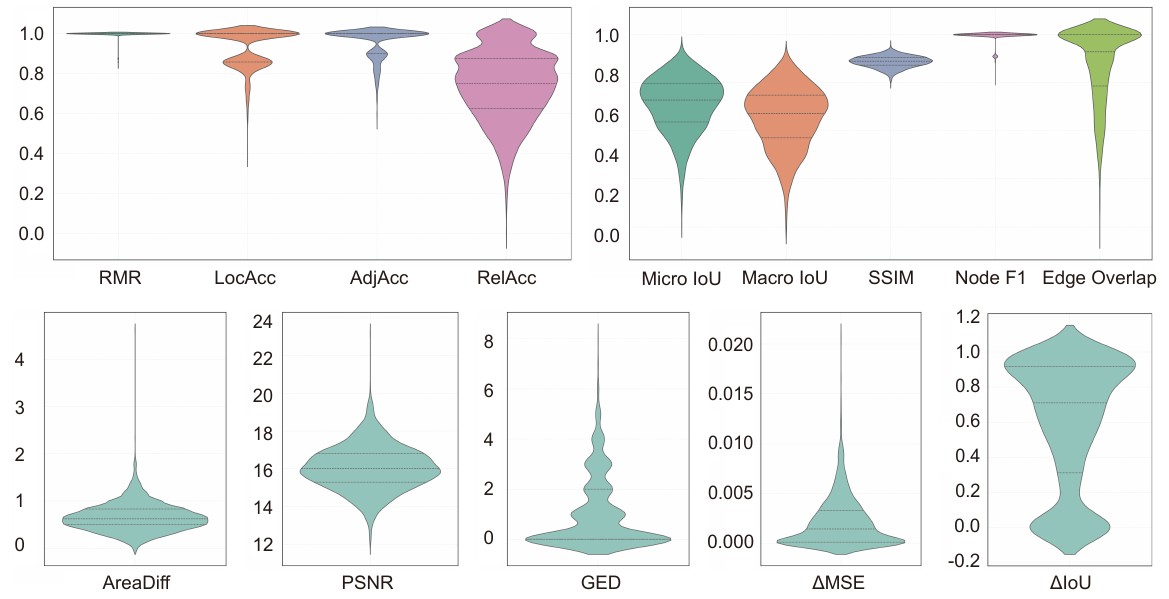}
    \caption{
    \textbf{Result distribution across tasks.}
    \textbf{HouseMind-O} maintains stable, high-performance distributions in understanding and generation,
    with the bimodal $\Delta\text{IoU}$ in editing mainly caused by unclear spatial instructions.
    }
    \label{fig:stability}
\end{figure*}

In the \textbf{understanding task}, \textbf{HouseMind-O} demonstrates accurate prediction of
room types, room locations, and room adjacency relations.
The Room Area Difference is generally controlled within $2\,\mathrm{m}^2$,
indicating stable quantitative reasoning of room size.
However, the prediction of spatial relations between rooms remains relatively weaker,
mainly due to minor inconsistencies in the definition of inter-room positional relationships
within the training dataset.

In the \textbf{generation task}, pixel-level metrics exhibit more uniform distributions
while maintaining consistently high values without significant outliers.
For structural metrics, most generated floor plans achieve $\text{Node~F1}=1$,
demonstrating nearly perfect alignment between predicted and target room types.
More than half of the samples also reach $\text{Edge~Overlap}=1$ and $\text{GED}=0$,
indicating precise prediction of room topological relations.
Only a few cases present slight deviations, which can be further improved in subsequent training stages.

In the \textbf{editing task}, the distribution of $\Delta\text{IoU}$ shows a distinct bimodal pattern.
This arises from a subset of editing prompts that do not explicitly specify
the insertion position of the new room,
leading to ambiguous geometric modifications.
When the editing instruction clearly describes both the target room location and size,
the predicted results are generally accurate and consistent with the ground truth.

\subsection{Pixel-Structure Coupling Analysis}

\noindent
We further analyze the relationship between pixel-level accuracy (Macro~IoU)
and structural consistency (Edge~Overlap) across three generation methods.

\begin{figure*}[t]
    \centering
    \includegraphics[width=\linewidth]{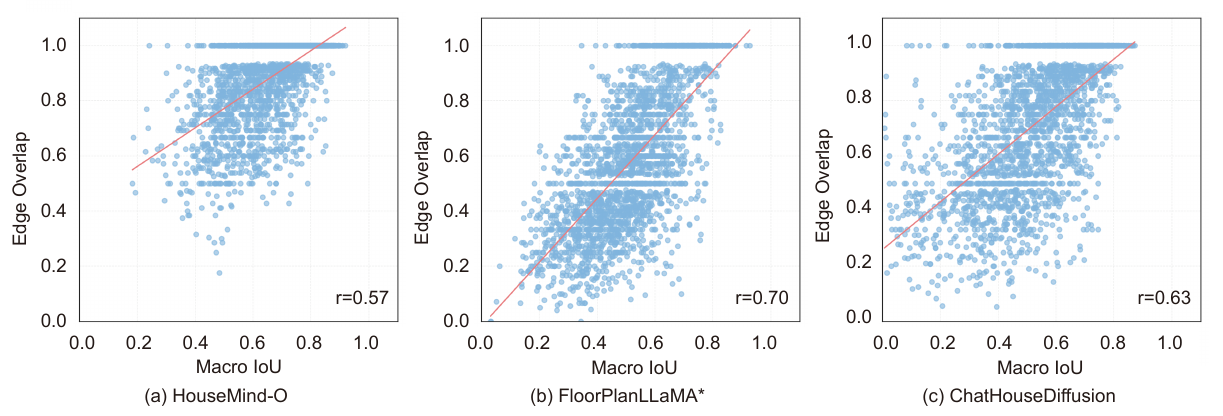}
    \caption{
    \textbf{Pixel–structure coupling analysis.}
    Scatter plots show the correlation between Macro~IoU and Edge~Overlap
    for three generation paradigms.
    }
    \label{fig:coupling}
\end{figure*}

As shown in Fig.~\ref{fig:coupling},
FloorPlanLLaMA~(our re-implementation), which encodes the entire layout as a single image sequence,
and ChatHouseDiffusion, which performs global image-to-layout diffusion generation,
both exhibit relatively strong correlations ($r=0.70$ and $r=0.63$, respectively)
between the two metrics.
This indicates that pixel-level alignment and structural accuracy are inherently interdependent in holistic generation.
As models pursue higher pixel fidelity, their topological precision also increases; however, this often comes at the cost of reduced flexibility in structural reasoning, as they overfit to visual similarity.

In contrast, \textbf{HouseMind-O}, equipped with room tokenization,
shows a lower correlation ($r=0.57$),
indicating that the pixel and topological aspects are partially decoupled.
Most data points lie in the upper region of the plot,
where structural consistency remains high, even when pixel IoU slightly decreases.
This reflects the model’s priority in maintaining accurate topological relations
across rooms, which better aligns with practical architectural requirements.

Overall, the reduced coupling in \textbf{HouseMind-O} demonstrates that
room-wise tokenization enables the model to reason about spatial topology
independently from pixel reconstruction,
achieving structure-aware generation that is both controllable and semantically consistent.

\subsection{Methodological Comparison and Discussion}

\noindent
Beyond quantitative metrics, we compare the underlying paradigms
adopted by representative model families to clarify the advantages of the proposed \textbf{HouseMind}.

General-purpose multimodal models such as GPT-5 integrate diffusion-based visual modules
for open-domain image generation and editing.
While these models excel in semantic reasoning and zero-shot transfer,
they lack geometric and topological priors, leading to spatially invalid or unstructured layouts.

Autoregressive text-to-layout generators like FloorPlanLLaMA
encode entire floor plans as sequential token streams.
This paradigm ensures high pixel fidelity and global semantic coherence
but suffers from limited controllability and no editing capability,
as the generation order constrains spatial reasoning.

Diffusion-based holistic generators, exemplified by ChatHouseDiffusion,
produce visually realistic and smooth layouts through iterative denoising.
However, their reliance on pixel-level reconstruction often causes inconsistent topology
and unstable room adjacency, which undermines functional correctness.

In contrast, the proposed \textbf{HouseMind} adopts a room-tokenized multimodal reasoning paradigm,
jointly modeling text, geometry, and topology within a unified LLM framework.
This design enables room-wise controllability and structural preservation,
achieving consistent topology even when visual pixel alignment fluctuates.
Unlike holistic approaches, \textbf{HouseMind} can independently reason about spatial structure
while maintaining semantic coherence across understanding, generation, and editing tasks.

Overall, this comparison highlights a paradigm shift from visually driven diffusion or autoregressive generation toward structure-aware multimodal reasoning, representing a critical step toward advancing controllable architectural layout generation.
\section{Prompts for Generation Cases}
\label{app:generation_prompts}

The following prompts correspond to the generation results 
illustrated in Fig. 4.
Each prompt describes the intended spatial relationships and functional layout 
used as textual input for the generation task.

\vspace{3pt}
\begin{tcolorbox}[promptstyle, title={Prompt G1}]
\footnotesize\ttfamily
\obeylines
The layout centers around a spacious Living Room, occupying the central area of the floor plan with an area of 26 square meters.
To the west of the Living Room lies the Master Room, which connects directly to it and also extends slightly above the Balcony situated to the south.
The Kitchen, positioned in the northwest corner, is smaller in size at 6 square meters and connects to both the Living Room on its right and the Bathroom directly below it.
The Bathroom, located just northwest of center and slightly west of the Living Room, is adjacent to both the Kitchen above and the Master Room below, forming a vertical sequence of rooms along the western side.
To the northeast, the Second Room sits comfortably next to the Living Room, sharing a direct right-of relationship with it.
Finally, the Balcony, positioned directly south of the Living Room, spans the southern edge of the plan and lies below both the Living Room and the Master Room, which projects slightly over it from the northwest.
All rooms are efficiently arranged, with clear adjacency and vertical alignment enhancing the flow between functional zones.
\end{tcolorbox}

\vspace{3pt}
\begin{tcolorbox}[promptstyle, title={Prompt G2}]
\footnotesize\ttfamily
\obeylines
Create a house layout that reflects the described spatial logic. The layout centers on a spacious LivingRoom, which is flanked to the left by the SecondRoom and to the right by the MasterRoom. Above the LivingRoom, toward the northeast, sit the Kitchen and Bathroom side by side, with the Kitchen positioned to the left of and slightly above the Bathroom. The Bathroom lies directly above the MasterRoom, while the Kitchen extends slightly northeastward, with a narrow Balcony positioned to its right and above the MasterRoom. A larger Balcony stretches along the southern edge of the LivingRoom, connecting to both the MasterRoom and Kitchen via diagonal relationships. The overall arrangement creates a functional, tiered flow from the central living area to private and service spaces.
\end{tcolorbox}

\vspace{3pt}
\begin{tcolorbox}[promptstyle, title={Prompt G3}]
\footnotesize\ttfamily
\obeylines
The layout centers on a spacious Living Room, occupying 36 square meters and positioned at the heart of the floor plan.
To the north, the Kitchen (4 sq. meters) sits directly above the Living Room and is connected to a smaller Balcony (2 sq. meters) that extends above it, with the Kitchen positioned just below this northern balcony.
To the northwest, the Second Room (10 sq. meters) lies above the Living Room and to the left of the Kitchen, with a diagonal connection beneath it to the northern balcony.
Directly south of the Living Room is the Master Room (8 sq. meters), which also connects to a larger Balcony (6 sq. meters) situated below it.
To the east of the Living Room lies the Bathroom (8 sq. meters), positioned to its right, with the Master Room lying diagonally below and to the left of it.
The two balconies—smaller to the north and larger to the south—frame the upper and lower edges of the layout, with the northern balcony adjacent to both the Kitchen and Second Room, and the southern balcony extending beneath the Master Room and aligned with the Living Room’s southern edge.
All rooms are arranged in a cohesive, vertically and horizontally connected sequence, with the Living Room serving as the central hub linking all other spaces.
\end{tcolorbox}

\section{Generalization and Real-World Deployment}
\label{sec:appendix-generalization}

This section provides a more detailed analysis of the generalization ability and robustness of \textbf{HouseMind} beyond the standard in-distribution evaluation.

\subsection{Modeling Design for Generalization}

From a modeling perspective,\textbf{ HouseMind }does not directly learn raw room distributions in pixel or coordinate space. Instead, spatial layouts are first discretized through a VQ-VAE encoder, which transforms continuous geometric configurations into structured spatial tokens. This representation abstracts away low-level coordinate noise and encourages the model to learn higher-level spatial relationships (e.g., adjacency, functional grouping, alignment patterns).

By operating in this tokenized latent space, the LLM captures relational and compositional regularities rather than memorizing specific layout instances. This representation-driven learning mechanism enables the model to generalize beyond layouts that strictly follow the empirical training distribution.

\subsection{Out-of-Distribution Evaluation}

To further evaluate generalization, we construct a deliberately more challenging out-of-distribution (OOD) test set for generation. Compared with the standard test split, this OOD set includes:

\begin{itemize}
    \item Studio layouts (no separate bedroom);
    \item Layouts with two living rooms;
    \item Layouts including uncommon room types (e.g., wine cellar treated as storage);
    \item Logically invalid configurations (e.g., centrally located balcony);
    \item Irregular building outlines;
    \item Prompts without explicitly specified room types.
\end{itemize}

\begin{figure}[ht]
  \vspace{-8pt}
  \centering
  \includegraphics[width=\linewidth]{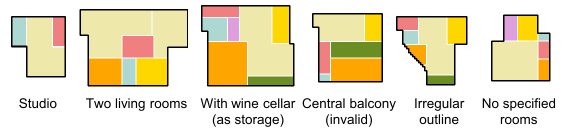}
  \vspace{-20pt}
   \caption{Results of \textbf{HouseMind} on out-of-distribution cases.}
   \label{fig:out-of-distribution}
   \vspace{-8pt}
\end{figure}

Representative examples are shown in Fig.~\ref{fig:out-of-distribution}.

Results indicate that \textbf{HouseMind} maintains structurally consistent generation across most uncommon layouts and room-type configurations. Even when room combinations deviate from typical training patterns, the model preserves global structural coherence and largely respects adjacency as well as functional grouping constraints. In addition, prompt paraphrasing does not significantly degrade structural validity, suggesting robustness to linguistic variations. Failure cases primarily arise under highly complex or mutually conflicting constraints. In such scenarios, the model tends to favor layouts that align with realistic architectural priors rather than strictly following logically inconsistent or physically implausible instructions (e.g., placing a balcony at the geometric center of the building). This behavior indicates that the model internalizes real-world spatial distributions and architectural regularities, which function as an implicit prior during generation.

\subsection{Robustness Enhancement in Deployment}

To further enhance robustness in real-world deployment, we implement several additional strategies in our public platform, available at \url{https://housemind.ai-structure.com/}:

\begin{enumerate}
    \item \textbf{Structured prompt guidance}: predefined room type selections and prompt templates are provided to reduce ambiguity and improve the clarity of user inputs.
    \item \textbf{Intermediate prompt reformulation}: an auxiliary LLM reformulates raw user instructions into structured representations that are better aligned with \textbf{HouseMind}'s token space.
    \item \textbf{Post-generation validation}: automatic validation of room types, counts, and adjacency relations is performed, with regeneration triggered when inconsistency with the input prompt is detected.
\end{enumerate}

These mechanisms collectively improve robustness under noisy, incomplete, or partially inconsistent user inputs, while preserving structural validity and generative diversity.

\end{document}